\documentclass[twocolumn]{autart}    

\usepackage{graphicx}          
\usepackage{amsmath,amssymb}
\usepackage{bm}
\usepackage{color}
\usepackage{subcaption}

\begin{document}

\begin{frontmatter}

\title{Lyapunov-Based Deep Learning Control\\ for Robots with Unknown Jacobian}





\author[Singapore]{Koji Matsuno}, 
\author[Singapore]{Chien Chern Cheah}


\address[Singapore]{School of Electrical and Electronic Engineering, Nanyang Technological University, 50 Nanyang Avenue, 639798, Singapore}
          
\begin{keyword}                           
Deep learning; robot control; stability.               
\end{keyword}                             

\begin{abstract}                          
Deep learning, with its exceptional learning capabilities and flexibility, has been widely applied in various applications. However, its black-box nature poses a significant challenge in real-time robotic applications, particularly in robot control, where trustworthiness and robustness are critical in ensuring safety. In robot motion control, it is essential to analyze and ensure system stability, necessitating the establishment of methodologies that address this need. This paper aims to develop a theoretical framework for end-to-end deep learning control that can be integrated into existing robot control theories. The proposed control algorithm leverages a modular learning approach to update the weights of all layers in real time, ensuring system stability based on Lyapunov-like analysis. Experimental results on industrial robots are presented to illustrate the performance of the proposed deep learning controller. The proposed method offers an effective solution to the black-box problem in deep learning, demonstrating the possibility of deploying real-time deep learning strategies for robot kinematic control in a stable manner. This achievement provides a critical foundation for future advancements in deep learning based real-time robotic applications.
\end{abstract}

\end{frontmatter}

\section{Introduction}
\hspace*{1em}Deep learning has played a pivotal role in the transformation of our society into a new era marked by advancements in artificial intelligence. Its impact is profound, notably elevating the capabilities of tasks like natural language processing and image recognition. This progress is fueled by harnessing vast datasets and tapping into high-performance computing resources. These advancements have captured global interest, with deep learning being applied in diverse fields, including robotics. In these applications, users often seek to understand the reasoning behind the conclusions made by neural networks to improve or enhance performance. Nevertheless, elucidating the inner workings of these computations poses a significant challenge. That is, understanding how the network derives outputs from inputs and what knowledge it acquires through learning remains unclear. This conundrum is commonly referred to as the black-box problem of deep learning. The integration of deep learning into robotics can introduce unforeseen behaviors, potentially compromising stability and hence safety. Given the critical importance of safety, trustworthiness, and robustness in robotics, addressing this challenge is paramount. Hence, a comprehensive theoretical understanding of deep learning becomes essential in ensuring the reliability and predictability of deep-learning based robotic systems \cite{sunderhauf2018limits}. \\
\hspace*{1em}In many modern robot systems, sensory feedback control is employed to perform a wide variety of tasks in unknown environments. However, stability issues may arise due to environmental factors, and it is crucial to achieve a precise understanding, prediction, and control of the robot's behavior. One way to ensure stability is to utilize model-based control theory, which mathematically represents the robot model to cancel out nonlinearity and derive a stable linear model. However, if there exists a significant modeling error, the control system cannot be assured to operate properly as the performance of the model-based control method relies on the accuracy of the model. Takegaki and Arimoto \cite{takegaki1981new} laid the foundation for the development of non-model-based robot control methods using the Lyapunov method. It was demonstrated that a simple joint-space PD controller with only gravity compensation could achieve global asymptotic stability for setpoint control tasks. For trajectory tracking control tasks with unknown dynamic parameters, Slotine and Li \cite{slotine1987adaptive} developed a globally convergent adaptive controller based on the idea of the sliding vector. To deal with robot control tasks with unknown kinematic parameters, Cheah et al. \cite{cheah2003approximate} \cite{cheah2006adaptive} developed approximate Jacobian and adaptive Jacobian task-space controllers. The stability analysis of the nonlinear closed-loop robot control systems was developed using Lyapunov analysis and since then, substantial advancements have been obtained in both stability analysis and controller design in robotics based on the Lyapunov method \cite{arimoto1996control} \cite{slotine1991applied} \cite{spong2022historical}. \\
\hspace*{1em}Because of their universal approximation properties for any compact set \cite{cybenko1989approximation} \cite{hecht1987kolmogorov}, neural networks are commonly used in scenarios where robot models are unknown. Nonetheless, their learning process typically relies on backpropagation, which may not guarantee system stability. To address this issue, the Lyapunov-like method has been employed in the development of control frameworks and the derivation of network weight update laws. Many researchers have contributed to the development of stable adaptive neural network controllers. In their pioneering work, Sanner and Slotine 
\cite{sanner1992gaussian} applied radial basis functions within a stable one-layer neural network, proving the system stability. Lewis et al.  \cite{lewis1996multilayer} demonstrated that a two-layer neural network could be updated in a stable manner and developed a controller for robots with unknown dynamics. For stable neural network control with unknown kinematic model, Lyu and Cheah \cite{lyu2020data}, and Yilmaz and Krstic \cite{yilmaz2023accelerated} introduced adaptive neural network Jacobian control methods. These methods utilize a single-layer neural network and analyze stability using the Lyapunov method. Although the stability of the neural network control systems can be analyzed, the results are limited to shallow networks with only one or two layers. \\
\hspace*{1em}In recent decades, deep neural networks have garnered significant attention because they surpass shallow networks in various tasks due to their advanced representational capabilities \cite{bengio2009learning} \cite{mhaskar2017and} \cite{mhaskar2016deep} \cite{montufar2014number}. However, despite the importance of their theoretical analysis, results on this topic are currently limited compared to those for shallow networks. Deriving stable update laws founded on theoretical analysis poses a significant challenge because deep neural networks iteratively compute weights within nonlinear activation functions. Nguyen and Cheah \cite{nguyen2022analytic} developed a framework where the weights are updated layer by layer to form a deep neural network. Therefore, this framework is only applicable to repetitive tasks where one layer is updated for each iteration. Sun et al. \cite{sun2021lyapunov} developed a multiple timescale learning architecture for nonlinear dynamic systems where the output layer weights are updated in real time. However, only the drift dynamics are approximated by deep neural networks, and it is assumed that the parameters of the control effectiveness matrix can be updated by traditional adaptive controller. To update all the weights, Le et al. \cite{le2021real} introduced constraints for the real-time adaptation of inner-layer weights but they did not offer detailed guidance on optimizing the design of these laws. Patil et al. \cite{patil2021lyapunov} also developed a recursive representation for the inner layers of the network. However, the method requires prior knowledge of the convex set, which is often not available under conditions of an uncertain or unknown system. In \cite{le2021real} and \cite{patil2021lyapunov}, the control effectiveness matrix of nonlinear systems is assumed to be known or can be simplified as an identity matrix. These studies \cite{le2021real} \cite{patil2021lyapunov} \cite{sun2021lyapunov} focus on nonlinear dynamic systems with unknown drift dynamics. However, in robotic systems, kinematic uncertainty is an inherent challenge within control systems. Li et al. \cite{li2023analytic} \cite{li2023theoretical} proposed an end-to-end learning algorithm that employs a virtual learning system and collaborative learning across multiple subsystems to update the weights of each layer simultaneously. Despite these efforts, it remains uncertain whether deep learning techniques can be effectively integrated with the latest advancements in Lyapunov-based robotic control theories. This uncertainty could either result in reinventing the field or underutilizing existing knowledge, potentially leading to inefficiencies in research progress. \\
\hspace*{1em}This paper proposes an end-to-end deep learning control architecture for robots with unknown kinematics. By adopting a modular learning approach, the stability of deep learning control systems can be analyzed using the Lyapunov-like method in time domain. To the best of our knowledge, this is the first result where a stable end-to-end deep-learning robot control problem has been formulated and solved to enable integration with existing robot control theories. This integration allows for the incorporation of state-of-the-art advancements in robotics, eliminating the need to reinvent existing technologies. The proposed neural network is structured to learn the input-output relationship in an end-to-end manner and is divided into two parts: the estimated-error based layers and the tracking-error based layers. The learning process involves training the model by comparing the output of each layer with the robot's current state in the estimated-error based layers, and with the target state in the tracking-error based layers. The weights across all layers are updated simultaneously using a modular approach to minimize the errors. To validate the effectiveness of the proposed method, the controller is implemented on an industrial robot to illustrate its performance. \\
\hspace*{1em}The remainder of this paper is organized as follows. Section 2 provides a detailed description of the proposed deep learning control architecture. Section 3 presents a stability analysis using the Lyapunov-like method. Section 4 presents the experimental setup and results, verifying the performance of the proposed method. Finally, section 5 concludes the paper by summarizing the key contributions and indicating the impact of our work on robot control using deep neural networks. 
\section{Deep Learning Based Kinematic Control}
\begin{figure*}[htbp]
    \centering
    \begin{subfigure}[b]{\textwidth}
        \centering
        \includegraphics[width=1\textwidth]{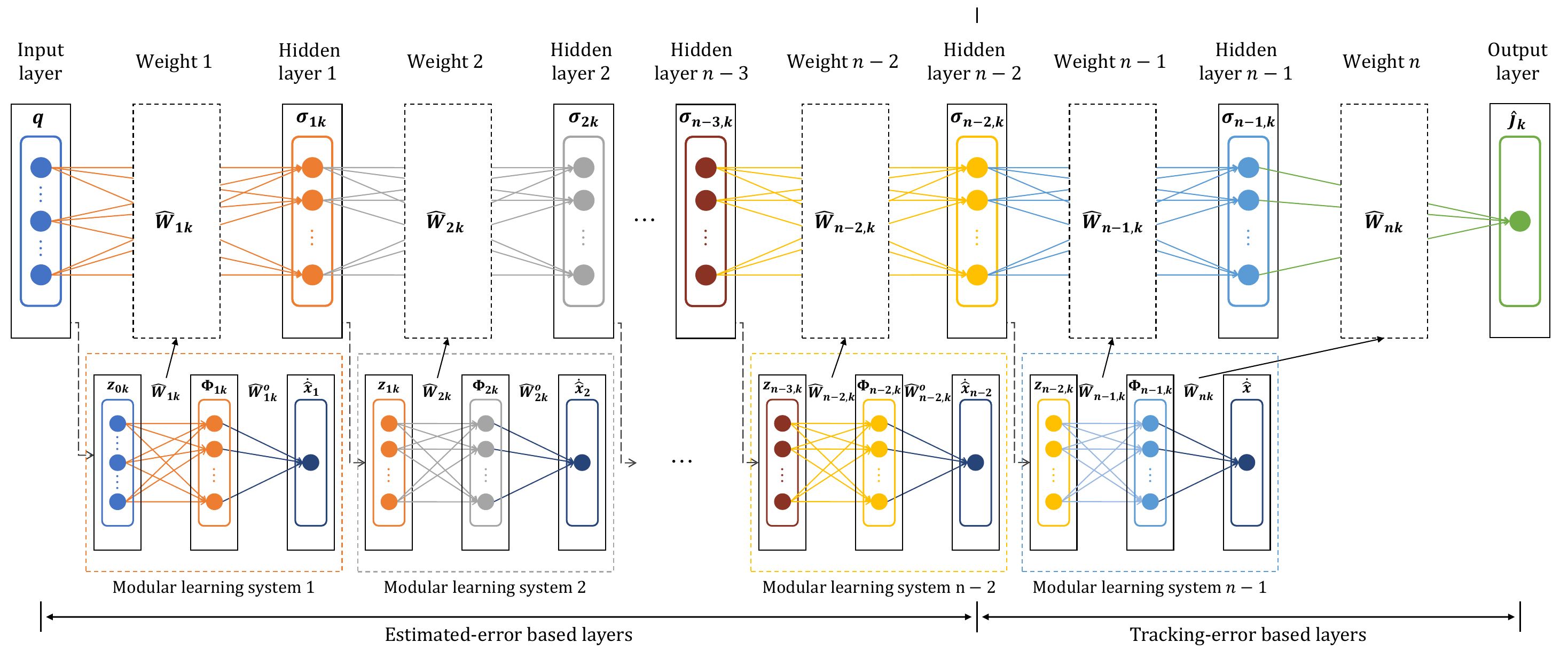}
        \caption{Building Block for Estimating each Column of the Jacobian Matrix}
        \label{fig:sub1}
    \end{subfigure}
    \par\vspace{0.5cm}
    \begin{subfigure}[b]{\textwidth}
        \centering
        \includegraphics[width=1\textwidth]{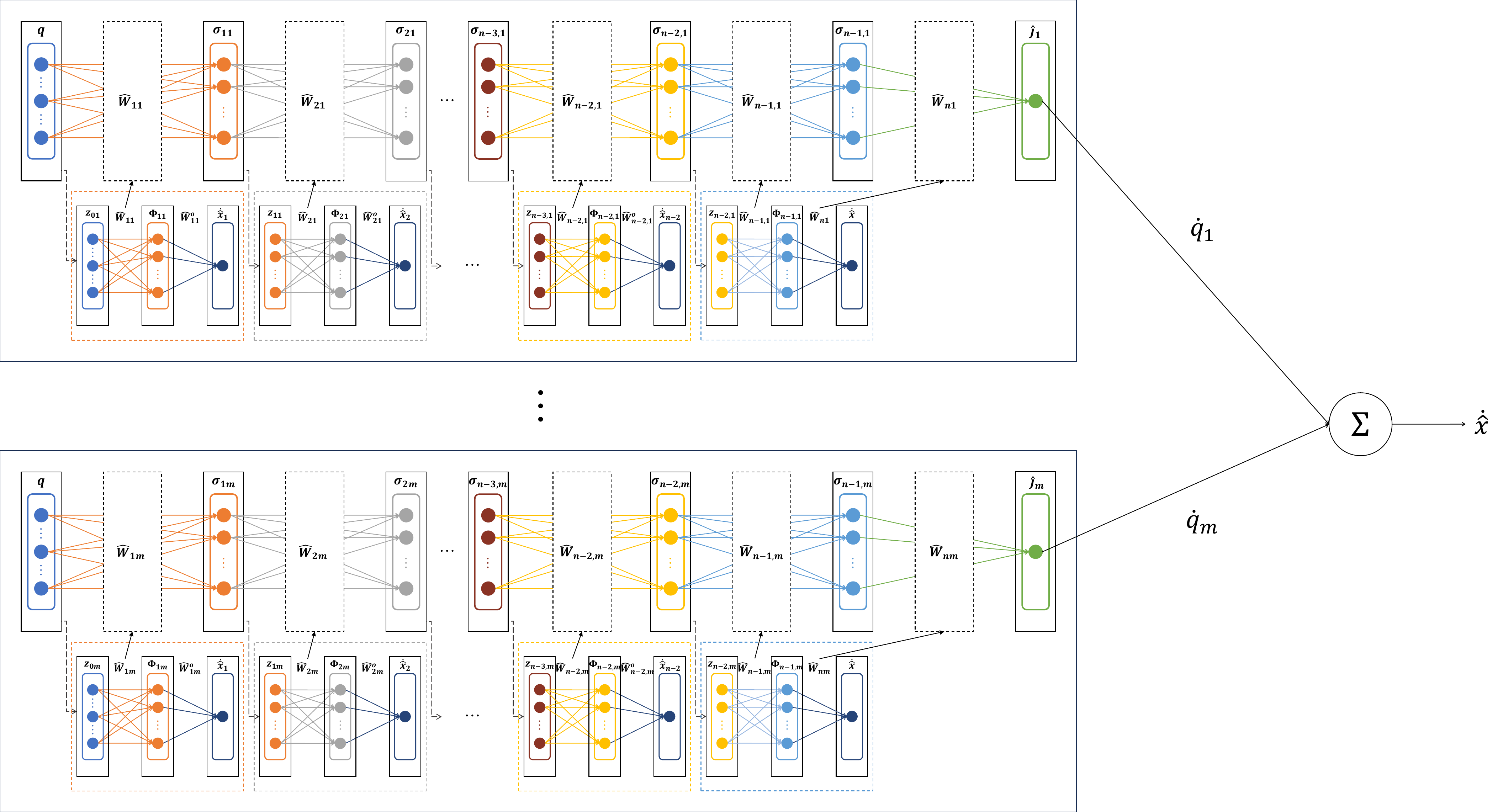}
        \caption{Overall Structure for Estimating the Jacobian Matrix}
        \label{fig:sub2}
    \end{subfigure}
    \caption{Neural Network Structure of End-to-End Deep Learning for the Approximate Jacobian}
    \label{fig:main}
\end{figure*}

\hspace*{1em}The kinematic relationship between the end-effector velocity vector $\dot{\bm x}$ and the joint velocity vector $\dot{\bm q}$ for a $m$-degree of freedom robot is given by
\begin{equation}
\label{eq00}
\begin{split}
\dot{\bm x} = \bm{J}\left(\bm q\right)\dot{\bm q}
\end{split}
\end{equation}
where $\bm{J}\left(\bm q\right)$ is the manipulator Jacobian matrix and $\bm{q}$ is the joint angle vector. Equation $(\ref{eq00})$ implies that the task-space velocity vector can be expressed as linear combinations of the columns of the Jacobian matrix multiplied by the elements of the joint-space velocity vector, that is
\begin{align}
\label{eq0}
\dot{\bm x} & = \bm{j}_1\left(\bm{q}\right)\dot q_1 + \bm{j}_2\left(\bm{q}\right)\dot q_2 + \cdots + \bm{j}_m\left(\bm{q}\right)\dot q_m \notag \\
&= \sum_{k=1}^{m}\bm{j}_k\left(\bm{q}\right)\dot q_k.
\end{align}
When the parameters and structure of the Jacobian are unknown, neural networks can be used to approximate them. To preserve the structure of the Jacobian matrix described by equation $(\ref{eq0})$, the structure of a $n$-layer neural network is designed as follows:
\begin{align}
\label{eq000}
\sum_{k=1}^{m}\bm{j}_k\left(\bm{q}\right)\dot q_k &= \sum_{k=1}^{m}\bm{W}_{nk}\bm\sigma_{n-1,k}\left(\bm{W}_{n-1,k}\bm\sigma_{n-2,k}\left(\cdots \right. \right. \notag \\
&\quad \left. \left. \bm\sigma_{2k}\left(\bm{W}_{2k}\bm\sigma_{1k}\left(\bm{W}_{1k}\bm q\right)\right)\right)\right)\dot q_k + \bm\epsilon_{nk}
\end{align}
where $\bm{W}_{lk}$ $\left(l = 1, 2, \ldots, n\right)$ and $\bm\sigma_{lk}$ $\left(l = 1, 2, \ldots, n-1\right)$ represent the ideal weight matrices and the activation function vectors of the $l$th hidden layer corresponding to the $k$th column of $\bm{J}\left(\bm q\right)$ respectively, and $\bm\epsilon_{nk}$ denotes the approximation errors. Since the ideal weight matrices are unknown, the Jacobian is estimated as
\begin{align}
\label{eq0000}
\sum_{k=1}^{m}\hat{\bm j}_k\left(\bm{q}\right)\dot q_k &= \sum_{k=1}^{m}\hat{\bm W}_{nk}\bm\sigma_{n-1,k}\left(\hat{\bm W}_{n-1,k}\bm\sigma_{n-2,k}\Big(\cdots \Big. \right. \notag \\
&\quad \Big. \left. \bm\sigma_{2k}\left(\hat{\bm W}_{2k}\bm\sigma_{1k}\left(\hat{\bm W}_{1k}\bm q\right)\right)\Big)\right)\dot q_k \notag \\
&\triangleq  \hat{\bm J}\left(\bm q, \hat{\bm W}_{\forall lk}\right)\dot{\bm q}
\end{align}
where $\hat{\bm W}_{lk}$ $\left(l = 1, 2, \ldots, n\right)$ represent the estimated weight matrices. Using the estimated Jacobian matrix, the kinematic control law is proposed as
\begin{align}
\label{eq16}
\dot{\bm q} &= \hat{\bm J}^{\dag}\left(\bm q, \hat{\bm W}_{\forall lk}\right)\dot{\bm x}_d - k_p\hat{\bm J}^{\mathrm T}\left(\bm q, \hat{\bm W}_{\forall lk}\right)\Delta\bm\xi \notag \\
&\quad - k_s\hat{\bm J}^{\dag}\left(\bm q, \hat{\bm W}_{\forall lk}\right)\mathrm {sat}\left(\Delta\bm x\right)
\end{align}
where $\hat{\bm J}^{\dag}\left(\bm q, \hat{\bm W}_{\forall lk}\right)$ is the pseudoinverse of $\hat{\bm J}\left(\bm q, \hat{\bm W}_{\forall lk}\right)$, $\bm{\dot x}_d$ is a desired velocity vector of the end-effector, $k_{p}$ and $k_{s}$ are positive gains, $\Delta\bm x = \bm x - \bm x_d$ is the tracking error vector defined as the difference between the actual position vector $\bm x$ and the desired position vector $\bm x_d$ with the elements $\Delta x_i = x_i - x_{di}$ for $i = 1, 2, \ldots, p$. The region error vector $\Delta\bm{\xi} = \left[\Delta\xi_1, \Delta\xi_2, \ldots, \Delta\xi_p\right]^\mathrm{T}$ is described by 
\begin{equation*}
\label{eq17}
\Delta{\xi}_i = \mathrm {max}\left(0, \frac{\Delta{x}_i^2}{b_i^2} - 1\right)^{N-1}\Delta{x}_i
\end{equation*}
where $b_i$ is a desired positive bound for $\Delta x_i$, and $N$ is a positive integer greater than or equal to 2. A smooth saturation function vector $\mathrm {sat}\left(\Delta\bm x\right) = \left[\mathrm {sat}\left(\Delta x_1\right), \mathrm {sat}\left(\Delta x_2\right), \cdots, \mathrm {sat}\left(\Delta x_p\right)\right]^\mathrm{T}$ is defined as
\begin{equation*}
\label{eq19}
\mathrm {sat}\left(\Delta{x}_i\right) = 
\begin{cases}
\sin\left(\frac{\pi}{2a_i}\Delta{x}_i\right) & \mathrm{if} \left|\Delta{x}_i\right| < a_i \\
\mathrm{sgn}\left(\Delta{x}_i\right) & \mathrm{otherwise}
\end{cases}
\end{equation*}
where ${a_i}$ are positive constants that are less than or equal to $b_i$. \\
\hspace*{1em}The proposed learning architecture of deep neural-network based Jacobian matrix comprises estimated-error based layers and tracking-error based layers as shown in Fig. \ref{fig:main}. The estimated-error based layer is iterated multiple times to form deeper layers, which culminate in the last two layers of the tracking-error based layers. The learning method propagates the input through each layer, while simultaneously updating the entire weights in a modular approach to minimize the error between the output of each layer and the corresponding target output. In the estimated-error based layers, each layer is computed using modular learning systems to obtain the outputs. These outputs are integrated to calculate the estimated position, which is then compared to the actual position to compute the estimated error. The weights are adjusted using the estimated-error based update laws to be described later. Each subsequent layer is trained by utilizing the hidden layer of the previous learning system as input, continuing this process until the last two layers. At this stage, the output weights are discarded, and only the input weights are used to approximate the Jacobian. Similarly, the tracking-error based layers also perform forward propagation calculations, and the weights are adjusted according to the tracking-error based update laws to be discussed later. \\
\hspace*{1em}Based on the proposed learning method described in Fig. \ref{fig:main}, the estimated task-space velocity vector for $l$th modular learning system in the estimated-error based layers are proposed as
\begin{align}
\label{eq1}
& \dot{\hat{\bm x}}_l = \sum_{k=1}^{m}\hat{\bm W}^o_{lk}\bm{\Phi}_{lk}\left(\hat{\bm W}_{lk}\bm{z}_{l-1,k}\right) \notag \\
& \phantom{\dot{\hat{\bm x}}_l =} + k_{pl}\Delta\hat{\bm \xi}_l + k_{sl}\mathrm {sat}\left(\Delta\hat{\bm x}_l\right), \\
\label{eq2}
& \bm{z}_{l-1,k} = \bm\sigma_{l-1,k}\left(\hat{\bm W}_{l-1,k}\bm\sigma_{l-2,k}\Big(\cdots \Big.\right. \notag \\
& \phantom{\bm{z}_{l-1,k} =} \Big.\left. \bm\sigma_{2k}\left(\hat{\bm W}_{2k}\bm\sigma_{1k}\left(\hat{\bm W}_{1k}\bm q\right)\right)\Big)\right)
\end{align}
where $\bm{\Phi}_{lk}\left(\cdot\right) = \bm\sigma_{lk}\left(\cdot\right)\dot q_k$, $\hat{\bm W}^o_{lk}$ and $\hat{\bm W}_{lk}$ are the estimated output weight matrices and input weight matrices respectively, and $k_{pl}$ and $k_{sl}$ are positive gains, $\bm{z}_{l-1,k}$ is treated as the input to the $l$th learning system with $\bm{z}_{0k} = \bm q$ for $l = 1$ and defined by equation $(\ref{eq2})$ for $l = 2, 3, \ldots, n-2$, $\Delta\bm{\hat x}_l = \bm x - \hat{\bm x}_l$ is the estimated position error vector with the estimated position vector as $\hat{\bm x}_l$, and the elements are $\Delta{\hat x}_{li} = x_{i} - {\hat x}_{li}$ for $i = 1, 2, \ldots, p$. The estimated region error vector $\Delta\bm{\hat\xi}_l = \left[\Delta{\hat\xi}_{l1}, \Delta{\hat\xi}_{l2}, \ldots, \Delta{\hat\xi}_{lp}\right]^\mathrm{T}$ is expressed as 
\begin{equation*}
\label{eq17}
\Delta{\hat\xi}_i = \mathrm {max}\left(0, \frac{\Delta{\hat x}_{li}^2}{b_{li}^2} - 1\right)^{N-1}\Delta{\hat x}_{li}
\end{equation*}
where $b_{li}$ are desired positive bounds for $\Delta{\hat x}_{li}$. A smooth saturation function vector for estimated-error based layers $\mathrm {sat}\left(\Delta\hat{\bm x}_l\right) = \left[\mathrm {sat}\left(\Delta{\hat x}_{l1}\right), \mathrm {sat}\left(\Delta{\hat x}_{l2}\right), \ldots, \mathrm {sat}\left(\Delta{\hat x}_{lp}\right)\right]^\mathrm{T}$ is given by
\begin{equation*}
\label{eq19}
\mathrm {sat}\left(\Delta{\hat x}_{li}\right) = 
\begin{cases}
\sin\left(\frac{\pi}{2a_{li}}\Delta{\hat x}_{li}\right) & \mathrm{if} \left|\Delta{\hat x}_{li}\right| < a_{li} \\
\mathrm{sgn}\left(\Delta{\hat x}_{li}\right) & \mathrm{otherwise}
\end{cases}
\end{equation*}
where $a_{li}$ are positive constants that are less than or equal to $b_{li}$. Equation $(\ref{eq1})$ serves as a velocity observer of each layer of the estimated-error based layers for the actual velocity. \\
\hspace*{1em}The weights of Jacobian for the estimated-error based layers are updated as
\begin{align}
\label{eq9}
& \dot{\hat{\bm W}}_{lk} = \alpha_{l}\hat{\bm\Phi}^{\prime\rm T}_{lk}\hat{\bm W}^{o\rm T}_{lk}\Delta\hat{\bm\xi}_l\bm{z}^{\rm T}_{l-1,k} - \alpha_{l}\beta_{l}\left|\left|\Delta\hat{\bm\xi}_l\right|\right|\hat{\bm W}_{lk}, \\
\label{eq10}
& \dot{\hat{\bm W}}_{lk}^o = \alpha^o_l\Delta\hat{\bm\xi}_l\left(\hat{\bm\Phi}^{\rm T}_{lk} - \bm{z}^{\rm T}_{l-1,k}\hat{\bm W}^{\rm T}_{lk}\hat{\bm \Phi}^{\prime\rm T}_{lk}\right) \notag \\
& \phantom{\dot{\hat{\bm W}}_{lk}^o=} - \alpha^o_l\beta^o_l\left|\left|\Delta\hat{\bm\xi}_l\right|\right|\hat{\bm W}^o_{lk}
\end{align}
where $\alpha_{l}$, $\alpha^o_l$,  $\beta_{l}$ and $\beta^o_l$ are positive constants, and $\hat{\bm\Phi}^{\prime}_{lk}$ are diagonal matrices with the elements of the gradients of $\hat{\bm\Phi}_{lk} = \bm{\Phi}_{lk}\left(\hat{\bm W}_{lk}\bm{z}_{l-1,k}\right)$ at $\hat{\bm W}_{lk}\bm{z}_{l-1,k}$. The weights of the tracking-error based layers are updated as
\begin{align}
\label{eq25}
& \dot{\hat{\bm W}}_{n-1,k} = \alpha_{n-1}\hat{\bm\Phi}^{\prime\rm T}_{n-1,k}\hat{\bm W}^{\rm T}_{nk}\Delta\bm\xi\bm{z}^{\rm T}_{n-2,k} \notag \\
& \phantom{\dot{\hat{\bm W}}_{n-1,k}=} - \alpha_{n-1}\beta_{n-1}\left|\left|\Delta\bm\xi\right|\right|\hat{\bm W}_{n-1,k}, \\
\label{eq261}
& \dot{\hat{\bm W}}_{nk} = \alpha_n\Delta\bm\xi\left(\hat{\bm\Phi}^{\rm T}_{n-1,k} - \bm{z}^{\rm T}_{n-2,k}\hat{\bm W}^{\rm T}_{n-1,k}\hat{\bm\Phi}^{\prime\rm T}_{n-1,k}\right) \notag \\
& \phantom{\dot{\hat{\bm W}}_{nk}=} - \alpha_n\beta_n\left|\left|\Delta\bm\xi\right|\right|\hat{\bm W}_{nk}
\end{align}
where $\alpha_{n-1}$, $\alpha_n$, $\beta_{n-1}$ and $\beta_n$ are positive constants, and $\hat{\bm\Phi}^{\prime}_{n-1,k}$ are diagonal matrices with the elements of the gradients of $\hat{\bm\Phi}_{n-1,k} = \bm{\Phi}_{n-1,k}\left(\hat{\bm W}_{n-1,k}\bm{z}_{n-2,k}\right)$ at $\hat{\bm W}_{n-1,k}\bm{z}_{n-2,k}$. 
\section{Stability Analysis}
\hspace*{1em}There exist an ideal weight matrix $\bm{W}_{lk}$, an ideal output weight matrix $\bm{W}^o_{lk}$, and an approximation error $\bm\epsilon_{lk}$ for each layer of the estimated-error based layers, such that the actual velocity vector can be expressed as
\begin{align}
\label{eq1100}
\dot{\bm x} = \sum_{k=1}^{m}\bm{W}^o_{lk}\bm{\Phi}_{lk}\left(\bm{W}_{lk}\bm{z}_{l-1,k}\right) + \bm\epsilon_{lk}.
\end{align}
Similarly, for the tracking-error based layers, the actual velocity vector can be represented by an ideal weight matrix $\bm{W}_{n-1k}$, an ideal output weight matrix $\bm{W}_{nk}$, and an approximation error $\bm\epsilon_{nk}$, formulated as
\begin{align}
\label{eq1101}
\dot{\bm x} = \sum_{k=1}^{m}\bm{W}_{nk}\bm{\Phi}_{n-1,k}\left(\bm{W}_{n-1,k}\bm{z}_{l-2,k}\right) + \bm\epsilon_{nk}.
\end{align}
\hspace*{1em}To show the stability analysis of the robot system, a Lyapunov-like function candidate is defined as
\begin{align}
\label{eq11}
V &= \sum_{l=1}^{n-2}P\left(\Delta\hat{\bm x}_l\right) + P\left(\Delta\bm{x}\right) \notag \\
&\quad + \sum_{l=1}^{n-2}\sum_{k=1}^{m}\frac{1}{2\alpha_{l}}\mathrm{tr}\left(\Delta\bm{W}^{\rm T}_{lk}\Delta\bm{W}_{lk}\right) \notag \\
&\quad + \sum_{l=1}^{n-2}\sum_{k=1}^{m}\frac{1}{2\alpha^o_l}\mathrm{tr}\left(\Delta\bm{W}^{o\rm T}_{lk}\Delta\bm{W}^o_{lk}\right) \notag \\
&\quad + \frac{1}{2\alpha_{n-1}}\sum_{k=1}^{m}\mathrm{tr}\left(\Delta\bm{W}^{\rm T}_{n-1,k}\Delta\bm{W}_{n-1,k}\right) \notag \\ 
&\quad + \frac{1}{2\alpha_n}\sum_{k=1}^{m}\mathrm{tr}\left(\Delta\bm{W}^{\rm T}_{nk}\Delta\bm{W}_{nk}\right)
\end{align}
where $\displaystyle P\left(\Delta\hat{\bm x}_l\right) = \sum_{i=1}^{p}\frac{1}{2N}\mathrm{max}\left(0, \frac{\Delta{\hat x}_{li}^2}{b_{li}^2} - 1\right)^{N}$ and $\displaystyle P\left(\Delta\bm{x}\right) = \sum_{i=1}^{p}\frac{1}{2N}\mathrm{max}\left(0, \frac{\Delta{x}_i^2}{b_i^2} - 1\right)^{N}$ represent the potential functions corresponding to the gradients of the region errors $\displaystyle\Delta\bm{\hat\xi}_l = \frac{\partial P\left(\Delta\hat{\bm x}_l\right)}{\partial \Delta\hat{\bm x}_l}$ and $\displaystyle\Delta\bm\xi = \frac{\partial P\left(\Delta\bm{x}\right)}{\partial \Delta\bm x}$, $\rm{tr}\left(\cdot\right)$ signifies the trace operator defined as the sum of the diagonal elements of a matrix, $\Delta\bm{W}_{lk} = \bm{W}_{lk} - \hat{\bm W}_{lk}$, $\Delta\bm{W}^o_{lk} = \bm{W}^o_{lk} - \hat{\bm W}^o_{lk}$, $\Delta\bm{W}_{n-1,k} = \bm{W}_{n-1,k} - \hat{\bm W}_{n-1,k}$ and $\Delta\bm{W}_{nk} = \bm{W}_{nk} - \hat{\bm W}_{nk}$ denote the weight estimation error matrices. Differentiating $V$ with respect to time, and substituting the update laws $(\ref{eq9})$, $(\ref{eq10})$, $(\ref{eq25})$ and $(\ref{eq261})$ into it yields
\begin{align}
\dot V &= \sum_{l=1}^{n-2}\Delta\hat{\bm\xi}^{\rm T}_l\Delta\dot{\hat{\bm x}}_l + \Delta\bm\xi^{\mathrm T}\Delta\dot{\bm x} \notag \\
&\quad - \sum_{l=1}^{n-2}\sum_{k=1}^{m}\mathrm{tr}\left(\Delta\bm{W}^{\rm T}_{lk}\hat{\bm\Phi}^{\prime\rm T}_{lk}\hat{\bm W}^{o\rm T}_{lk}\Delta\hat{\bm \xi}_l\bm{z}^{\rm T}_{l-1,k}\right) \notag \\
&\quad + \sum_{l=1}^{n-2}\sum_{k=1}^{m}\beta_{l}\left|\left|\Delta\hat{\bm \xi}_l\right|\right|\mathrm{tr}\left(\Delta\bm{W}^{\rm T}_{lk}\hat{\bm W}_{lk}\right) \notag \\
&\quad - \sum_{l=1}^{n-2}\sum_{k=1}^{m}\mathrm{tr}\left(\Delta\bm{W}^{o\rm T}_{lk}\Delta\hat{\bm\xi}_l\left(\hat{\bm\Phi}^{\rm T}_{lk} 
- \bm{z}^{\rm T}_{l-1,k}\hat{\bm W}^{\rm T}_{lk}\hat{\bm\Phi}^{\prime}_{lk}\right)\right) \notag 
\end{align}
\begin{align}
\label{eq122}
&\quad + \sum_{l=1}^{n-2}\sum_{k=1}^{m}\beta^o_l\left|\left|\Delta\hat{\bm\xi}_l\right|\right|\mathrm{tr}\left(\Delta\bm{W}^{o\rm T}_{lk}\hat{\bm W}^o_{lk}\right) \notag \\
&\quad - \sum_{k=1}^{m}\mathrm{tr}\left(\Delta\bm{W}^{\rm T}_{n-1,k}\hat{\bm\Phi}^{\prime\rm T}_{n-1,k}\hat{\bm W}^{\rm T}_{nk}\Delta\bm\xi\bm{z}^{\rm T}_{n-2,k}\right) \notag \\
&\quad + \beta_{n-1}\left|\left|\Delta\bm\xi\right|\right|\sum_{k=1}^{m}\mathrm{tr}\left(\Delta\bm{W}^{\rm T}_{n-1,k}\hat{\bm W}_{n-1,k}\right) \notag \\
&\quad - \sum_{k=1}^{m}\mathrm{tr}\left(\Delta\bm{W}^{\rm T}_{nk}\Delta\bm\xi\left(\hat{\bm\Phi}^{\rm T}_{n-1,k} 
 - \bm{z}^{\rm T}_{n-2,k}\hat{\bm W}^{\rm T}_{n-1,k}\hat{\bm\Phi}^{\prime}_{n-1,k}\right)\right) \notag \\
&\quad + \beta_n\left|\left|\Delta\bm\xi\right|\right|\sum_{k=1}^{m}\mathrm{tr}\left(\Delta\bm{W}^{\rm T}_{nk}\hat{\bm W}_{nk}\right) 
\end{align}
where $\Delta\dot{\hat{\bm x}}_l = \dot{\bm x} - \dot{\hat{\bm x}}_l$ represent the velocity estimatied error vector at $l$th layer, and $\Delta{\dot{\bm x}} = \dot{\bm x} - \dot{\bm x}_d$ denotes the velocity tracking error vector. \\
\hspace*{1em}To proceed, the velocity estimated error vector in the multilayer neural network is represented as
\begin{align}
\label{eq4}
\sum_{l=1}^{n-2}\Delta\dot{\hat{\bm x}}_l &= \sum_{l=1}^{n-2}\sum_{k=1}^{m}\bm{W}^o_{lk}\Delta\bm{\Phi}_{l-1,k} \notag \\
&\quad + \sum_{l=1}^{n-2}\sum_{k=1}^{m}\Delta\bm{W}^o_{lk}\hat{\bm\Phi}_{l-1,k} \notag \\
&\quad + \sum_{l=1}^{n-2}\sum_{k=1}^{m}\bm\epsilon_{lk} \notag \\
&\quad  - \sum_{l=1}^{n-2}k_{pl}\Delta\hat{\bm\xi}_l - \sum_{l=1}^{n-2}k_{sl}\mathrm {sat}\left(\Delta\hat{\bm x}_l\right)
\end{align}
where $\Delta\bm{\Phi}_{lk} = \bm{\Phi}_{lk} - \hat{\bm\Phi}_{lk}$ with $\bm{\Phi}_{lk} = \bm{\Phi}_{lk}\left(\bm{W}_{lk}\bm{z}_{l-1,k}\right)$. Using the Taylor series expansion of $\hat{\bm\Phi}_{lk}$ about $\hat{\bm W}_{lk} = \bm{W}_{lk}$, $\Delta\bm{\Phi}_{lk}$ can be expressed as
\begin{align}
\label{eq5}
\Delta\bm{\Phi}_{lk} &= \bm{\Phi}^{\prime}_{lk}\Delta\bm{W}_{lk}\bm{z}_{l-1,k} + O\left(\left(\Delta\bm{W}_{lk}\bm{z}_{l-1,k}\right)^2\right) \notag \\
&= \hat{\bm\Phi}^{\prime}_{lk}\Delta\bm{W}_{lk}\bm{z}_{l-1,k} + \Delta\bm{\Phi}^{\prime}_{lk}\Delta\bm{W}_{lk}\bm{z}_{l-1,k} \notag \\
&\quad + O\left(\left(\Delta\bm{W}_{lk}\bm{z}_{l-1,k}\right)^2\right)
\end{align}
where $O\left(\left(\Delta\bm{W}_{lk}\bm{z}_{l-1,k}\right)^2\right)$ are the second and higher order terms. Since the activation functions (e.g., sigmoid and tanh function) and their derivative are bounded, the higher order terms in equation $(\ref{eq5})$ are bounded by 
\begin{align}
\label{eq6}
&\left|\left|\Delta\bm{\Phi}^{\prime}_{lk}\Delta\bm{W}_{lk}\bm{z}_{l-1,k} + O\left(\left(\Delta\bm{W}_{lk}\bm{z}_{l-1,k}\right)^2\right)\right|\right| \notag \\
&= \left|\left|\Delta\bm{\Phi}_{lk} - \hat{\bm\Phi}^{\prime}_{lk}\Delta\bm{W}_{lk}\bm{z}_{l-1,k}\right|\right| \notag \\
&\leq \left|\left|\Delta\bm{\Phi}_{lk}\right|\right| + \left|\left|\hat{\bm\Phi}^{\prime}_{lk}\Delta\bm{W}_{lk}\bm{z}_{l-1,k}\right|\right| \notag \\
&\leq c_{1k} + c_{2k}\left|\left|\Delta\bm{W}_{lk}\right|\right|
\end{align}
where $c_{1k}$ and $c_{2k}$ are positive constants. Substituting equation $(\ref{eq5})$ into equation $(\ref{eq4})$, the cumulative velocity estimated error vectors can be rewritten as
\begin{align}
\label{eq71}
\sum_{l=1}^{n-2}\Delta\dot{\hat{\bm x}}_l &= \sum_{l=1}^{n-2}\sum_{k=1}^{m}\hat{\bm W}^o_{lk}\hat{\bm\Phi}^{\prime}_{lk}\Delta\bm{W}_{lk}\bm{z}_{l-1,k} \notag \\
&\quad + \sum_{l=1}^{n-2}\sum_{k=1}^{m}\Delta\bm{W}^o_{lk}\left(\hat{\bm\Phi}_{lk} - \hat{\bm\Phi}^{\prime}_{lk}\hat{\bm W}_{lk}\bm{z}_{l-1,k}\right) \notag \\
&\quad - \sum_{l=1}^{n-2}k_{pl}\Delta\hat{\bm\xi}_l - \sum_{l=1}^{n-2}k_{sl}\mathrm {sat}\left(\Delta\hat{\bm x}_l\right) \notag \\
&\quad + \sum_{l=1}^{n-2}\sum_{k=1}^{m}\bm{d}_{lk}
\end{align}
where 
\begin{equation*}
\begin{split}
\label{eq7}
\bm{d}_{lk} &= \Delta\bm{W}^o_{lk}\hat{\bm\Phi}^{\prime}_{lk}\bm{W}_{lk}\bm{z}_{l-1,k} + \bm{W}^o_{lk}\Delta\bm{\Phi}^{\prime}_{lk}\Delta\bm{W}_{lk}\bm{z}_{l-1,k} \\
&\quad + \bm{W}^o_{lk}O\left(\left(\Delta\bm{W}_{lk}\bm{z}_{l-1,k}\right)^2\right) + \bm\epsilon_{lk}.
\end{split}
\end{equation*}
From inequality $(\ref{eq6})$, $\bm{d}_{lk}$ are also bounded by
\begin{align}
\label{eq8}
\left|\left|\bm{d}_{lk}\right|\right| &\leq c_{3k}\left|\left|\Delta\bm{W}^o_{lk}\right|\right|\left|\left|\bm{W}_{lk}\right|\right| \notag \\
&\quad + \left|\left|\bm{W}^o_{lk}\right|\right|\left(c_{1k} + c_{2k}\left|\left|\Delta\bm{W}_{lk}\right|\right|\right) + \left|\left|\bm\epsilon_{lk}\right|\right| \notag \\
&\leq C_{1k} + C_{2k}\left|\left|\Delta\bm{W}_{lk}\right|\right| + C_{3k}\left|\left|\Delta\bm{W}^o_{lk}\right|\right|
\end{align}
where $c_{3k}$ and $C_{ik}$ $\left(i = 1, 2, 3\right) $ are positive constants. \\ 
\hspace*{1em}For tracking-error based layers, there exists an estimated weight matrix $\hat{\bm W}_{n-1,k}$, and an estimated output weight matrix $\hat{\bm W}_{nk}$, such that the estimated velocity vector can be expressed as
\begin{align}
\label{eq11011}
\dot{\hat{\bm x}} = \sum_{k=1}^{m}\hat{\bm W}_{nk}\bm{\Phi}_{n-1,k}\left(\hat{\bm W}_{n-1,k}\bm{z}_{l-2,k}\right).
\end{align}
By substituting equation $(\ref{eq16})$ into equation $(\ref{eq0000})$, and rearranging it with equation $(\ref{eq11011})$, the velocity tracking error vector can be derived as
\begin{align}
\label{eq20}
\Delta\dot{\bm x} &= \bm{W}_{nk}\Delta\bm{\Phi}_{n-1,k} + \Delta\bm{W}_{nk}\hat{\bm\Phi}_{n-1,k} + \bm\epsilon_{nk} \notag \\
&\quad - k_p\hat{\bm J}\left(\bm q, \hat{\bm W}_{\forall lk}\right)\hat{\bm J}^{\mathrm T}\left(\bm q, \hat{\bm W}_{\forall lk}\right)\Delta\bm\xi \notag \\
&\quad - k_s\mathrm {sat}\left(\Delta\bm x\right)
\end{align}
where $\Delta\bm{\Phi}_{n-1,k} = \bm{\Phi}_{n-1,k} - \hat{\bm\Phi}_{n-1,k}$ with $\bm{\Phi}_{n-1,k} = \bm{\Phi}_{n-1,k}\left(\bm{W}_{n-1,k}\bm{z}_{n-2,k}\right)$. Expanding $\hat{\bm\Phi}_{n-1,k}$ around $\hat{\bm W}_{n-1,k} = \bm{W}_{n-1,k}$ in the Taylor series, $\Delta\bm{\Phi}_{n-1,k}$ can be represented as
\begin{align}
\label{eq21}
\Delta\bm{\Phi}_{n-1,k} &= \bm{\Phi}^{\prime}_{n-1,k}\Delta\bm{W}_{n-1,k}\bm{z}_{n-2,k} \notag \\
&\quad + O\left(\left(\Delta\bm{W}_{n-1,k}\bm{z}_{n-2,k}\right)^2\right) \notag \\
&= \hat{\bm\Phi}^{\prime}_{n-1,k}\Delta\bm{W}_{n-1,k}\bm{z}_{n-2,k} \notag \\
&\quad + \Delta\bm{\Phi}^{\prime}_{n-1,k}\Delta\bm{W}_{n-1,k}\bm{z}_{n-2,k} \notag \\
&\quad + O\left(\left(\Delta\bm{W}_{n-1,k}\bm{z}_{n-2,k}\right)^2\right) 
\end{align}
where
where $O\left(\left(\Delta\bm{W}_{n-1,k}\bm{z}_{n-2,k}\right)^2\right)$ is the second and higher order terms. Based on inequality $(\ref{eq6})$, the higher order terms in equation $(\ref{eq21})$ are bounded by
\begin{align}
\label{eq22}
&\left|\left|\Delta\bm{\Phi}^{\prime}_{n-1,k}\Delta\bm{W}_{n-1,k}\bm{z}_{n-2,k} + O\left(\left(\Delta\bm{W}_{n-1,k}\bm{z}_{n-2,k}\right)^2\right)\right|\right| \notag \\
&= \left|\left|\Delta\bm{\Phi}_{n-1,k} - \hat{\bm\Phi}^{\prime}_{n-1,k}\Delta\bm{W}_{n-1,k}\bm{z}_{n-2,k}\right|\right| \notag \\
&\leq \left|\left|\Delta\bm{\Phi}_{n-1,k}\right|\right| + \left|\left|\hat{\bm\Phi}^{\prime}_{n-1,k}\Delta\bm{W}_{n-1,k}\bm{z}_{n-2,k}\right|\right| \notag \\
&\leq c_{6k} + c_{7k}\left|\left|\Delta\bm{W}_{n-1,k}\right|\right|
\end{align}
where $c_{6k}$ and $c_{7k}$ are positive constants. Substituting equation $(\ref{eq21})$ into equation $(\ref{eq20})$, the velocity tracking error vector becomes
\begin{align}
\label{eq231}
\Delta\dot{\bm x} &= \sum_{k=1}^{m}\hat{\bm W}_{nk}\hat{\bm\Phi}^{\prime}_{n-1,k}\Delta\bm{W}_{n-1,k}\bm{z}_{n-2,k} \notag \\
&\quad + \sum_{k=1}^{m}\Delta\bm{W}_{nk}\left(\hat{\bm\Phi}_{n-1,k} - \hat{\bm\Phi}^{\prime}_{n-1,k}\hat{\bm W}_{n-1,k}\bm{z}_{n-2,k}\right) \notag \\
&\quad - k_p\hat{\bm J}\left(\bm q, \hat{\bm W}_{\forall lk}\right)\hat{\bm J}^{\mathrm T}\left(\bm q, \hat{\bm W}_{\forall lk}\right)\Delta\bm\xi \notag \\
&\quad - k_s\mathrm {sat}\left(\Delta\bm x\right) + \sum_{k=1}^{m}\bm{d}_{nk}
\end{align}
where
\begin{equation*}
\begin{split}
\label{eq233}
\bm{d}_{nk} &= \Delta\bm{W}_{nk}\hat{\bm\Phi}^{\prime}_{n-1,k}\bm{W}_{n-1,k}\bm{z}_{n-2,k} \\
&\quad + \bm{W}_{nk}\Delta\bm{\Phi}^{\prime}_{n-1,k}\Delta\bm{W}_{n-1,k}\bm{z}_{n-2,k} \\
&\quad + \bm{W}_{nk}O\left(\left(\Delta \bm{W}_{n-1,k}\bm{z}_{n-2,k}\right)^2\right) + \bm\epsilon_{nk}.
\end{split}
\end{equation*}
Using inequality $(\ref{eq6})$, $\bm{d}_{nk}$ is bounded by
\begin{align}
\label{eq24}
\left|\left|\bm{d}_{nk}\right|\right| &\leq c_{8k}\left|\left|\Delta\bm{W}_{nk}\right|\right|\left|\left|\bm{W}_{n-1,k}\right|\right| \notag \\
&\quad + \left|\left|\bm{W}_{nk}\right|\right|\left(c_{6k} + c_{7k}\left|\left|\Delta\bm{W}_{n-1,k}\right|\right|\right) + \left|\left|\bm\epsilon_{nk}\right|\right| \notag \\
&\leq C_{6k} + C_{7k}\left|\left|\Delta\bm{W}_{n-1,k}\right|\right| + C_{8k}\left|\left|\Delta\bm{W}_{nk}\right|\right|
\end{align}
where $c_{8k}$ and $C_{ik}$ $\left(i = 6, 7, 8\right)$ are positive constants. \\
\hspace*{1em}Taking the inner product of equation $(\ref{eq71})$ with $\Delta\bm{\hat\xi}$ and of equation $(\ref{eq231})$ with $\Delta\bm{\xi}$ provides
\begin{align}
\label{eq1111}
& \sum_{l=1}^{n-2}\Delta\hat{\bm\xi}^{\rm T}_l\Delta\dot{\hat{\bm x}}_l = \sum_{l=1}^{n-2}\sum_{k=1}^{m}\Delta\hat{\bm\xi}^{\rm T}_l\hat {\bm W}^o_{lk}\hat{\bm\Phi}^{\prime}_{lk}\Delta\bm{W}_{lk}\bm{z}_{l-1,k} \notag \\
& \phantom{\sum_{l=1}^{n-2}\Delta\hat{\bm\xi}^{\rm T}_l\Delta\dot{\hat{\bm x}}_l =} + \sum_{l=1}^{n-2}\sum_{k=1}^{m}\Delta\hat{\bm\xi}^{\rm T}_l\Delta\bm{W}^o_{lk}\left(\hat{\bm\Phi}_{lk} \right. \notag \\
& \phantom{\sum_{l=1}^{n-2}\Delta\hat{\bm\xi}^{\rm T}_l\Delta\dot{\hat{\bm x}}_l =} \left. - \hat{\bm\Phi}^{\prime}_{lk}\hat{\bm W}_{lk}\bm{z}_{l-1,k}\right) \notag \\
& \phantom{\sum_{l=1}^{n-2}\Delta\hat{\bm\xi}^{\rm T}_l\Delta\dot{\hat{\bm x}}_l =} - \sum_{l=1}^{n-2}k_{pl}\Delta\hat{\bm\xi}^{\rm T}_l\Delta\hat{\bm\xi}_l \notag \\
& \phantom{\sum_{l=1}^{n-2}\Delta\hat{\bm\xi}^{\rm T}_l\Delta\dot{\hat{\bm x}}_l =} - \sum_{l=1}^{n-2}k_{sl}\Delta\hat{\bm\xi}^{\rm T}_l\mathrm {sat}\left(\Delta\hat{\bm x}_l\right) \notag \\
& \phantom{\sum_{l=1}^{n-2}\Delta\hat{\bm\xi}^{\rm T}_l\Delta\dot{\hat{\bm x}}_l =} + \sum_{l=1}^{n-2}\sum_{k=1}^{m}\Delta\hat{\bm\xi}^{\rm T}_l\bm{d}_{lk}, \\
\label{eq1112}
& \Delta\bm\xi^{\mathrm T}\Delta\dot{\bm x} = \sum_{k=1}^{m}\Delta\bm\xi^{\mathrm T}\hat{\bm W}_{nk}\hat{\bm\Phi}^{\prime}_{n-1,k}\Delta\bm{W}_{n-1,k}\bm{z}_{n-2,k} \notag \\
& \phantom{\Delta\bm\xi^{\mathrm T}\Delta\dot{\bm x} =} + \sum_{k=1}^{m}\Delta\bm\xi^{\mathrm T}\Delta\bm{W}_{nk}\left(\hat{\bm\Phi}_{n-1,k} \right. \notag \\
& \phantom{\Delta\bm\xi^{\mathrm T}\Delta\dot{\bm x} =} \left. - \hat{\bm\Phi}^{\prime}_{n-1,k}\hat{\bm W}_{n-1,k}\bm{z}_{n-2,k}\right) \notag \\
& \phantom{\Delta\bm\xi^{\mathrm T}\Delta\dot{\bm x} =} - k_p\Delta\bm\xi^{\mathrm T}\hat{\bm J}\left(\bm q, \hat{\bm W}_{\forall lk}\right)\hat{\bm J}^{\mathrm T}\left(\bm q, \hat{\bm W}_{\forall lk}\right)\Delta\bm\xi \notag \\
& \phantom{\Delta\bm\xi^{\mathrm T}\Delta\bm{\dot x} =} - k_s\Delta\bm\xi^{\mathrm T}\mathrm {sat}\left(\Delta\bm x\right) + \sum_{k=1}^{m}\Delta\bm\xi^{\mathrm T}\bm{d}_{nk}.
\end{align}
Then, substituting these results into equation $(\ref{eq122})$ gives
\begin{align}
\label{eq12}
\dot V &= - \sum_{l=1}^{n-2}k_{pl}\Delta\hat{\bm\xi}^{\rm T}_l\Delta\hat{\bm\xi}_l \notag \\
&\quad - \sum_{l=1}^{n-2}k_{sl}\Delta\hat{\bm\xi}^{\rm T}_l\mathrm{sat}\left(\Delta\hat{\bm x}_l\right) + \sum_{l=1}^{n-2}\sum_{k=1}^{m}\Delta\hat{\bm\xi}^{\rm T}_l\bm{d}_{lk} \notag \\
&\quad + \sum_{l=1}^{n-2}\sum_{k=1}^{m}\beta_{l}\left|\left|\Delta\hat{\bm\xi}_l\right|\right|\mathrm{tr}\left(\Delta\bm{W}^{\rm T}_{lk}\left(\bm{W}_{lk} - \Delta\bm{W}_{lk}\right)\right) \notag \\
&\quad + \sum_{l=1}^{n-2}\sum_{k=1}^{m}\beta^o_l\left|\left|\Delta\hat{\bm\xi}_l\right|\right|\mathrm{tr}\left(\Delta\bm{W}^{o\rm T}_{lk}\left(\bm{W}^o_{lk} - \Delta\bm{W}^o_{lk}\right)\right) \notag \\
&\quad - k_p\Delta\bm\xi^{\mathrm T}\hat{\bm J}\left(\bm q, \hat{\bm W}_{\forall lk}\right)\hat{\bm J}^{\mathrm T}\left(\bm q, \hat{\bm W}_{\forall lk}\right)\Delta\bm\xi \notag \\
&\quad - k_s\Delta\bm\xi^{\mathrm T}\mathrm {sat}\left(\Delta\bm{x}\right) + \sum_{k=1}^{m}\Delta\bm\xi^{\mathrm T}\bm{d}_{nk} \notag \\
&\quad + \beta_{n-1}\left|\left|\Delta\bm\xi\right|\right|\sum_{k=1}^{m}\mathrm{tr}\left(\Delta\bm{W}^{\rm T}_{n-1,k}\left(\bm{W}_{n-1,k} - \Delta\bm{W}_{n-1,k}\right)\right) \notag \\
&\quad + \beta_n\left|\left|\Delta\bm\xi\right|\right|\sum_{k=1}^{m}\mathrm{tr}\left(\Delta\bm{W}^{\rm T}_{nk}\left(\bm{W}_{nk} - \Delta\bm{W}_{nk}\right)\right). 
\end{align}
\hspace*{1em}When $\left|\Delta\hat{\bm x}_l\right| \geq b_l$ and $\left|\Delta\bm{x}\right| \geq b$ are met, $\mathrm{sgn}\left(\Delta\hat{\bm x}_l\right) - \mathrm{sat}\left(\Delta\hat{\bm x}_l\right) = 0$ and $\mathrm{sgn}\left(\Delta\bm x\right) - \mathrm{sat}\left(\Delta\bm x\right) = 0$ hold. Conversely, when $\left|\Delta\hat{\bm x}_l\right| < b_l$ and $\left|\Delta\bm{x}\right| < b$, it follows that $\Delta\bm{\hat\xi} = 0$ and $\Delta\bm{\xi} = 0$. Therefore, 
\begin{align}
\label{eqe1}
&k_{sl}\Delta\hat{\bm\xi}^{\rm T}_l\mathrm{sat}\left(\Delta\hat{\bm x}_l\right) \notag \\
&= k_{sl}\Delta\hat{\bm\xi}^{\rm T}_l\mathrm{sgn}\left(\Delta\hat{\bm x}_l\right) - k_{sl}\Delta\hat{\bm\xi}^{\rm T}_l\left(\mathrm{sgn}\left(\Delta\hat{\bm x}_l\right) - \mathrm{sat}\left(\Delta\hat{\bm x}_l\right)\right) \notag \\
&\geq k_{sl}\left|\left|\Delta\hat{\bm\xi}_l\right|\right|
\end{align}
and 
\begin{align}
\label{eqe2}
&k_s\Delta\bm\xi^{\rm T}\mathrm{sat}\left(\Delta\bm x\right) \notag \\
&= k_s\Delta\bm\xi^{\rm T}\mathrm{sgn}\left(\Delta\bm x\right) - k_s\Delta\bm\xi^{\rm T}\left(\mathrm{sgn}\left(\Delta\bm x\right) - \mathrm{sat}\left(\Delta\bm x\right)\right) \notag \\
&\geq k_s\left|\left|\Delta\bm\xi\right|\right|
\end{align}
are satisfied. \\
\hspace*{1em}By applying inequalities $(\ref{eq8})$, $(\ref{eq24})$, $(\ref{eqe1})$ and $(\ref{eqe2})$ to equation $(\ref{eq12})$, $\dot V$ can be reformulated as
\begin{align}
\dot V &\leq - \sum_{l=1}^{n-2}k_{pl}\left|\left|\Delta\bm{\hat\xi}_l\right|\right|^2 - \sum_{l=1}^{n-2}k_{sl}\left|\left|\Delta\hat{\bm\xi}_l\right|\right| \notag \\
&\quad + \sum_{l=1}^{n-2}\sum_{k=1}^{m}\left|\left|\Delta\hat{\bm\xi}_l\right|\right|\left(C_{1k} + C_{2k}\left|\left|\Delta\bm{W}_{lk}\right|\right| + C_{3k}\left|\left|\Delta\bm{W}^o_{lk}\right|\right|\right) \notag \\
&\quad + \sum_{l=1}^{n-2}\sum_{k=1}^{m}\beta_{l}\left|\left|\Delta\hat{\bm\xi}_l\right|\right|\left|\left|\Delta\bm{W}_{lk}\right|\right|\left(C_{4k} - \left|\left|\Delta\bm{W}_{lk}\right|\right|\right) \notag \\
&\quad + \sum_{l=1}^{n-2}\sum_{k=1}^{m}\beta^o_l\left|\left|\Delta\hat{\bm\xi}_l\right|\right|\left|\left|\Delta\bm{W}^o_{lk}\right|\right|\left(C_{5k} - \left|\left|\Delta\bm{W}^o_{lk}\right|\right|\right) \notag \\
&\quad - k_p\left|\left|\hat{\bm J}^{\mathrm T}\left(\bm q, \hat{\bm W}_{\forall lk}\right)\Delta\bm\xi\right|\right|^2 - k_s\left|\left|\Delta\bm\xi\right|\right| \notag \\
&\quad + \sum_{k=1}^{m}\left|\left|\Delta\bm\xi\right|\right|\left(C_{6k} + C_{7k}\left|\left|\Delta\bm{W}_{n-1,k}\right|\right| + C_{8k}\left|\left|\Delta\bm{W}_{nk}\right|\right|\right) \notag \\
&\quad + \beta_{n-1}\left|\left|\Delta\bm\xi\right|\right|\sum_{k=1}^{m}\left|\left|\Delta\bm{W}_{n-1,k}\right|\right|\left(C_{9k} - \left|\left|\Delta\bm{W}_{n-1,k}\right|\right|\right) \notag \\
&\quad + \beta_n\left|\left|\Delta\bm\xi\right|\right|\sum_{k=1}^{m}\left|\left|\Delta\bm{W}_{nk}\right|\right|\left(C_{10k} - \left|\left|\Delta\bm{W}_{nk}\right|\right|\right) \notag \\
&\leq - \sum_{l=1}^{n-2}k_{pl}\left|\left|\Delta\hat{\bm\xi}_l\right|\right|^2 - \sum_{l=1}^{n-2}\left|\left|\Delta\hat{\bm\xi}_l\right|\right|\Biggl\{k_{sl} - \sum_{k=1}^{m}\Biggl[C_{1k} \Biggr.\Biggr. \notag \\
&\quad \Biggl.\Biggl. + \left|\left|\Delta\bm{W}_{lk}\right|\right|\left(C_{2k} + \beta_{l}C_{4k} - \beta_{l}\left|\left|\Delta\bm{W}_{lk}\right|\right|\right) \Biggr.\Biggr. \notag \\
&\quad \Biggl.\Biggl. + \left|\left|\Delta\bm{W}^o_{lk}\right|\right|\left(C_{3k} + \beta^o_lC_{5k} - \beta^o_l\left|\left|\Delta\bm{W}^o_{lk}\right|\right|\right)\Biggr]\Biggr\} \notag \\
&\quad - k_p\left|\left|\hat{\bm J}^{\mathrm T}\left(\bm q, \hat{\bm W}_{\forall lk}\right)\Delta\bm\xi\right|\right|^2 - \left|\left|\Delta\bm\xi\right|\right|\Biggl\{k_s - \sum_{k=1}^{m}\Biggl[C_{6k} \Biggr.\Biggr. \notag
\end{align}
\begin{align}
\label{eq29}
&\quad \Biggl.\Biggl. + \left|\left|\Delta\bm{W}_{n-1,k}\right|\right|\left(C_{7k} + \beta_{n-1}C_{9k} - \beta_{n-1}\left|\left|\Delta\bm{W}_{n-1,k}\right|\right|\right) \Biggr.\Biggr. \notag \\
&\quad \Biggl.\Biggl. + \left|\left|\Delta\bm{W}_{nk}\right|\right|\left(C_{8k} + \beta_nC_{10k} - \beta_n\left|\left|\Delta\bm{W}_{nk}\right|\right|\right)\Biggr]\Biggr\}
\end{align}
where $C_{4k}$, $C_{5k}$, $C_{9k}$ and $C_{10k}$ are positive constants. To ensure $\dot V \leq 0$, the positive gains $k_{sl}$ and $k_s$ are selected to satisfy the conditions
\begin{align}
\label{eq13}
& k_{sl} 
\geq \sum_{k=1}^{m}C_{1k} + \sum_{l=1}^{n-2}\sum_{k=1}^{m}\beta_{l}\left|\left|\Delta\bm{W}_{lk}\right|\right|\left(C_{2,4k} -\left|\left|\Delta\bm{W}_{lk}\right|\right|\right) \notag \\
& \phantom{k_{sl} \geq} + \sum_{l=1}^{n-2}\sum_{k=1}^{m}\beta^o_l\left|\left|\Delta\bm{W}^o_{lk}\right|\right|\left(C_{3,5k} - \left|\left|\Delta\bm{W}^o_{lk}\right|\right|\right) \notag \\
& \phantom{k_{sl}} = \sum_{k=1}^{m}C_{1k} - \sum_{l=1}^{n-2}\sum_{k=1}^{m}\beta_{l}\left(\left|\left|\Delta\bm{W}_{lk}\right|\right| - \frac{C_{2,4k}}{2}\right)^2  \notag \\
& \phantom{k_{sl} =} - \sum_{l=1}^{n-2}\sum_{k=1}^{m}\beta^o_l\left(\left|\left|\Delta\bm{W}^o_{lk}\right|\right| - \frac{C_{3,5k}}{2}\right)^2 \notag \\
& \phantom{k_{sl} =} + \sum_{l=1}^{n-2}\sum_{k=1}^{m}\beta_{l}\frac{C^2_{2,4k}}{4} + \sum_{l=1}^{n-2}\sum_{k=1}^{m}\beta^o_l\frac{C^2_{3,5k}}{4}, \\
\label{eq31}
& k_s 
\geq \sum_{k=1}^{m}C_{6k} + \beta_{n-1}\sum_{k=1}^{m}\left|\left|\Delta\bm{W}_{n-1,k}\right|\right|\left(C_{7,9k} - \left|\left|\Delta\bm{W}_{n-1,k}\right|\right|\right) \notag \\
& \phantom{k_{s} \geq} + \beta_n\sum_{k=1}^{m}\left|\left|\Delta\bm{W}_{nk}\right|\right|\left(C_{8,10k} - \left|\left|\Delta\bm{W}_{nk}\right|\right|\right) \notag \\
& \phantom{k_{s}} = \sum_{k=1}^{m}C_{6k} - \beta_{n-1}\sum_{k=1}^{m}\left(\left|\left|\Delta\bm{W}_{n-1,k}\right|\right| - \frac{C_{7,9k}}{2}\right)^2 \notag \\
& \phantom{k_{s} =} - \beta_n\sum_{k=1}^{m}\left(\left|\left|\Delta\bm{W}_{nk}\right|\right| - \frac{C_{8,10k}}{2}\right)^2 \notag \\
& \phantom{k_{s} =} + \beta_{n-1}\sum_{k=1}^{m}\frac{C^2_{7,9k}}{4} + \beta_n\sum_{k=1}^{m}\frac{C^2_{8,10k}}{4}
\end{align}
where  $C_{2,4k} = \frac{C_{2k}}{\beta_{l}} + C_{4k}$, $C_{3,5k} = \frac{C_{3k}}{\beta^o_l} + C_{5k}$, $C_{7,9k} = \frac{C_{7k}}{\beta_{n-1}} + C_{9k}$, and $C_{8,10k} = \frac{C_{8k}}{\beta_n} + C_{10k}$. It is sufficient to choose
\begin{align}
\label{eq13}
& k_{sl} > \sum_{k=1}^{m}C_{1k} + \sum_{l=1}^{n-2}\sum_{k=1}^{m}\beta_{l}\frac{C^2_{2,4k}}{4} + \sum_{l=1}^{n-2}\sum_{k=1}^{m}\beta^o_l\frac{C^2_{3,5k}}{4}, \\
\label{eq31}
& k_{s} > \sum_{k=1}^{m}C_{6k} + \beta_{n-1}\sum_{k=1}^{m}\frac{C^2_{7,9k}}{4} + \beta_n\sum_{k=1}^{m}\frac{C^2_{8,10k}}{4}
\end{align}
which simplifies the original criteria while still guaranteeing that $\dot V \leq 0$. 
\newpage
\begin{thm}
The weight update laws given in equations $(\ref{eq9})$ to $(\ref{eq261})$ guarantee the convergence of the region errors, i.e., $\Delta\hat{\bm\xi}_l \rightarrow 0$ and $\Delta\bm\xi \rightarrow 0$ as $t \rightarrow \infty$.
\end{thm}
\vspace{-0.7cm}
\begin{pf}
The Lyapunov function $V$ in equation $(\ref{eq11})$ is positive definite. Under the conditions specified in equation $(\ref{eq13})$ and $(\ref{eq31})$, its time derivative satisfies $\dot{V} \leq 0$, ensuring that $V$ is non-increasing and bounded. According to equation $(\ref{eq122})$, the boundedness of the region errors $\Delta \hat{\bm\xi}_l$ and $\Delta \bm{\xi}$ implies that both the estimated errors $\Delta \bm{\hat{x}}_l$ and the tracking error $\Delta \bm{x}$ remain bounded. Additionally, the time derivatives $\Delta\dot{\hat{\bm x}}_l$ and $\Delta \dot{\bm x}$ are bounded, which leads to the boundedness of $\Delta\dot{\hat{\bm\xi}}_l$ and $\Delta\dot{\bm\xi}$. By applying Barbalat's lemma, $\Delta \hat{{\bm\xi}}_l \to 0$ and $\Delta \bm{\xi} \to 0$ as $t \to \infty$. 
\end{pf}
\begin{rem}
If the Jacobian matrix is known, the first and second terms in equation $(\ref{eq16})$ represent the traditional task-space control laws commonly used in robot kinematic control. Early works on visual servoing \cite{hutchinson1996tutorial} also assumed the Jacobian matrix to be exactly known. To address kinematic uncertainty, several adaptive Jacobian control laws have been developed \cite{cheah2015task}\cite{cheah2006adaptive}\cite{wang2016adaptive}, but in these cases, the structure of the Jacobian matrix is presumed to be known, with only the kinematic parameters updated online. To develop controllers when the Jacobian matrix is unknown, adaptive neural-network controllers based on shallow networks have been employed. In these approaches, the stability of control systems with uncertain kinematics can be analyzed using the Lyapunov method. However, deep neural networks introduce significant nonlinearity due to the use of nonlinear activation functions across multiple layers, making stability analysis, and particularly integration with existing theories, more challenging. This theoretical study provides the first insight into integrating deep neural network control with established robot control theories, potentially broadening the scope of research and applications for real-time deep learning in robotics.
\end{rem}
\begin{rem}
The region error $\Delta\bm\xi$ can be replaced by the tracking error $\Delta\bm x$. Consequently, the control law in equation $(\ref{eq16})$ is reformulated with the second term depending on $\Delta\bm x$. This adjustment also entails substituting the saturation function with the sign function, as indicated by
\begin{align}
\label{eq16a}
\dot{\bm q} &= \hat{\bm J}^{\dag}\left(\bm q, \hat{\bm W}_{\forall lk}\right)\dot{\bm x}_d - k_p\hat{\bm J}^{\mathrm T}\left(\bm q, \hat{\bm W}_{\forall lk}\right)\Delta\bm x \notag \\
&\quad - k_s\hat{\bm J}^{\dag}\left(\bm q, \hat{\bm W}_{\forall lk}\right)\mathrm {sgn}\left(\Delta\bm x\right).
\end{align}
Similarly, the estimated task-space velocity vector in equation $(\ref{eq1})$ is refined by replacing the estimated region error $\Delta\hat{\bm\xi}_l$ with the estimated error $\Delta\hat{\bm x}_l$ and substituting the saturation function with the sign function, as shown in \vspace{-0.1cm} 
\begin{align}
\label{eq1a}
& \dot{\hat{\bm x}}_l = \sum_{k=1}^{m}\hat{\bm W}^o_{lk}\bm{\Phi}_{lk}\left(\hat{\bm W}_{lk}\bm{z}_{l-1,k}\right) \notag \\
& \phantom{\dot{\hat{\bm x}}_l =} + k_{pl}\Delta\hat{\bm x}_l + k_{sl}\mathrm {sgn}\left(\Delta\hat{\bm x}_l\right).
\end{align}
These modifications also apply to the update laws, which are adjusted to incorporate the estimated and tracking errors as follows: 
\begin{align}
\label{eq9a}
& \dot{\hat{\bm W}}_{lk} = \alpha_{l}\hat{\bm\Phi}^{\prime\rm T}_{lk}\hat{\bm W}^{o\rm T}_{lk}\Delta\hat{\bm x}_l\bm{z}^{\rm T}_{l-1,k} - \alpha_{l}\beta_{l}\left|\left|\Delta\hat{\bm x}_l\right|\right|\hat{\bm W}_{lk}, \\
\label{eq10a}
& \dot{\hat{\bm W}}_{lk}^o = \alpha^o_l\Delta\hat{\bm x}_l\left(\hat{\bm\Phi}^{\rm T}_{lk} - \bm{z}^{\rm T}_{l-1,k}\hat{\bm W}^{\rm T}_{lk}\hat{\bm\Phi}^{\prime\rm T}_{lk}\right) \notag \\
& \phantom{\dot{\hat{\bm W}}_{lk}^o=} - \alpha^o_l\beta^o_l\left|\left|\Delta\hat{\bm x}_l\right|\right|\hat{\bm W}^o_{lk}, \\
\label{eq25a}
& \dot{\hat{\bm W}}_{n-1,k} = \alpha_{n-1}\hat{\bm\Phi}^{\prime\rm T}_{n-1,k}\hat{\bm W}^{\rm T}_{nk}\Delta\bm x\bm{z}^{\rm T}_{n-2,k} \notag \\
& \phantom{\dot{\hat{\bm W}}_{n-1,k}=} - \alpha_{n-1}\beta_{n-1}\left|\left|\Delta\bm x\right|\right|\hat{\bm W}_{n-1,k}, \\
\label{eq261a}
& \dot{\hat{\bm W}}_{nk} = \alpha_n\Delta\bm x\left(\hat{\bm\Phi}^{\rm T}_{n-1,k} - \bm{z}^{\rm T}_{n-2,k}\hat{\bm W}^{\rm T}_{n-1,k}\hat{\bm\Phi}^{\prime\rm T}_{n-1,k}\right) \notag \\
& \phantom{\dot{\hat{\bm W}}_{nk}=} - \alpha_n\beta_n\left|\left|\Delta\bm x\right|\right|\hat{\bm W}_{nk}.
\end{align}
Additionally, the potential functions $P(\Delta\hat{\bm x}_l)$ and $P(\Delta\bm{x})$ are revised to reflect the estimated and tracking errors, as given by
\begin{align}
\label{eq9ab}
& P\left(\Delta\hat{\bm x}_l\right) = \sum_{i=1}^{p}\frac{1}{2}\Delta{\hat x}_{li}^{2}, \\
\label{eq10ab}
& P\left(\Delta\bm{x}\right) = \sum_{i=1}^{p}\frac{1}{2}\Delta{x}_i^{2}.
\end{align}
It is important to note that these substitutions do not affect the fundamental results of the stability analysis. The changes only modify the form of the control law and update laws, while the overall conclusions regarding system stability remain the same. If the neural network approximation errors are negligible, the positive gains $k_{sl}$ and $k_s$ can be set to zero. 
\end{rem}
\section{Experiment}
\begin{figure*}[htbp]
    \centering
    \begin{subfigure}[t]{0.47\textwidth}
        \centering
        \includegraphics[width=\textwidth]{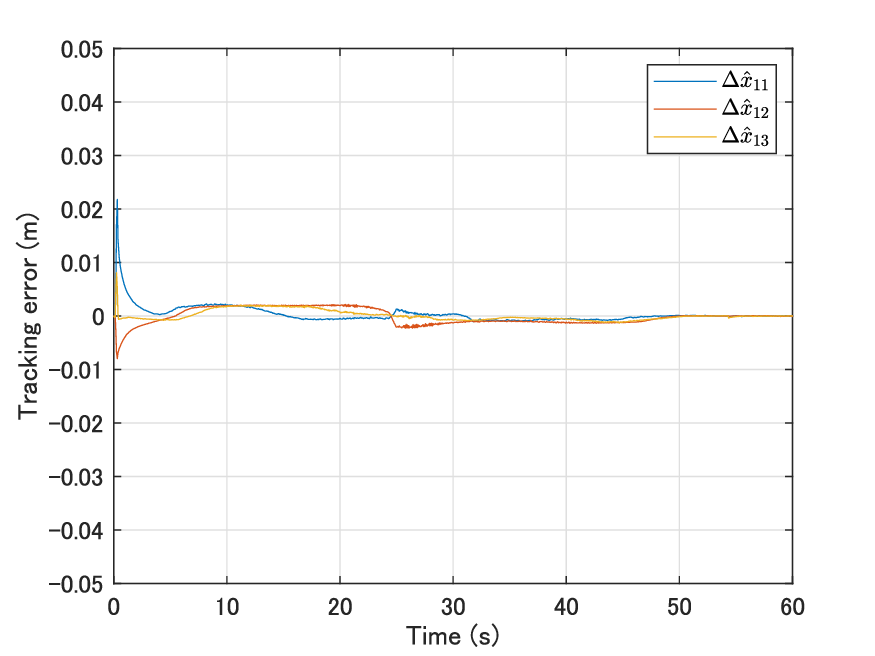}
        \caption{Estimated error for modular learning system 1}
        \label{fig:sub11}
    \end{subfigure}
    \hfill
    \begin{subfigure}[t]{0.47\textwidth}
        \centering
        \includegraphics[width=\textwidth]{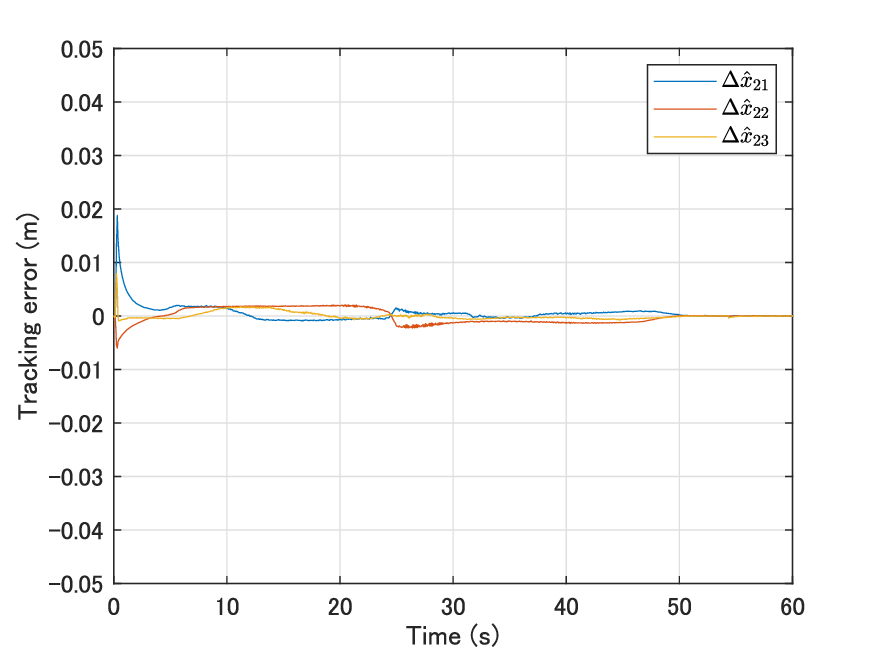}
        \caption{Estimated error for modular learning system 2}
        \label{fig:sub21}
    \end{subfigure}
    \vspace{1em}
    
    \begin{subfigure}[t]{0.47\textwidth}
        \centering
        \includegraphics[width=\textwidth]{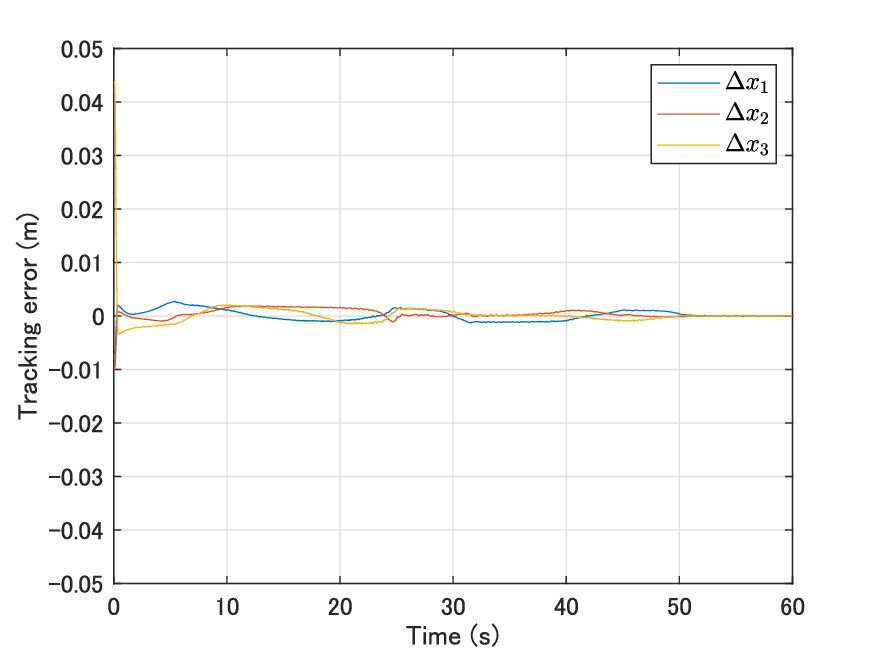}
        \caption{Tracking error}
        \label{fig:sub31}
    \end{subfigure}
    \hfill
    \begin{subfigure}[t]{0.47\textwidth}
        \centering
        \includegraphics[width=\textwidth]{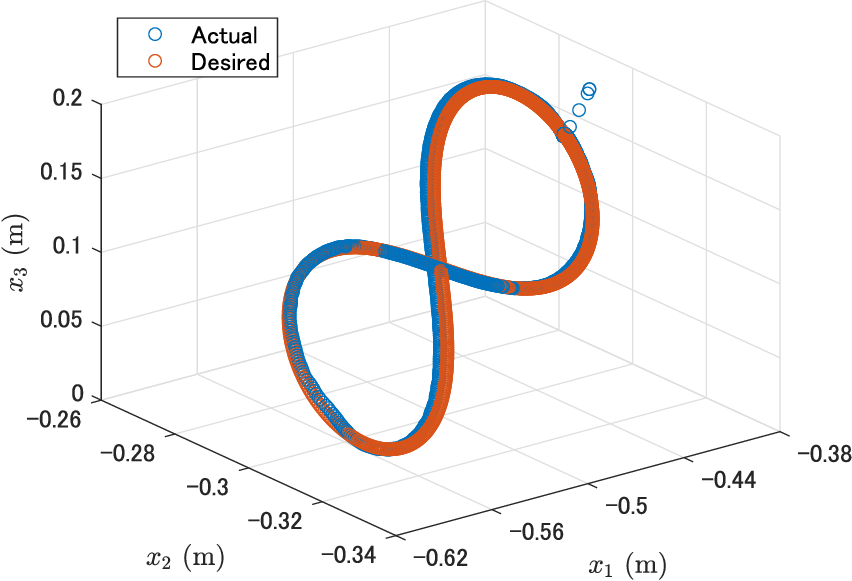}
        \caption{Trajectory tracking in 3D space}
        \label{fig:sub41}
    \end{subfigure}
    
    \caption{Results of region control}
    \label{fig:main2}
\end{figure*}

\begin{figure*}[htbp]
    \centering
    
    \begin{subfigure}[t]{0.47\textwidth}
        \centering
        \includegraphics[width=\textwidth]{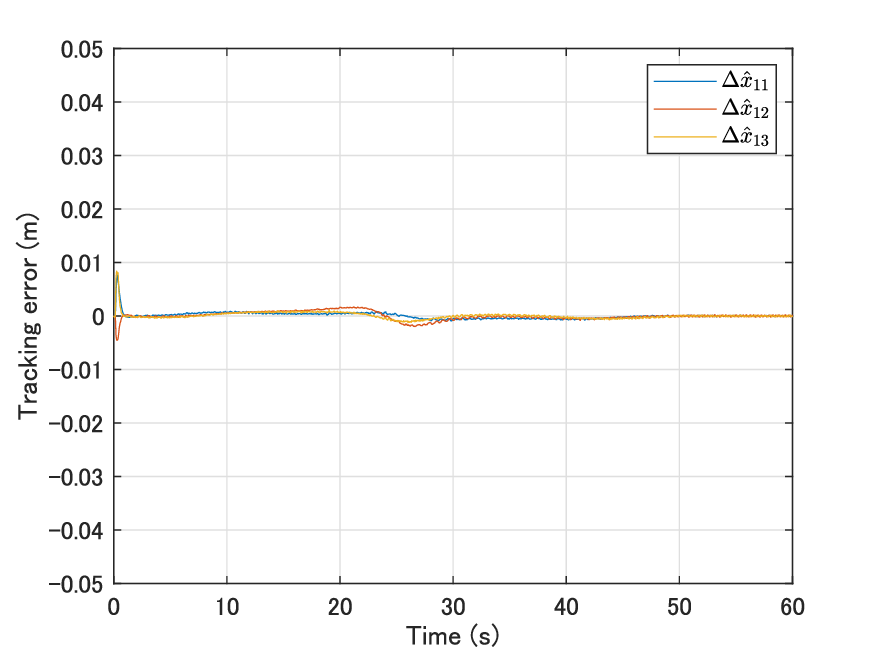}
        \caption{Estimated error for modular learning system 1}
        \label{fig:sub11}
    \end{subfigure}
    \hfill
    \begin{subfigure}[t]{0.47\textwidth}
        \centering
        \includegraphics[width=\textwidth]{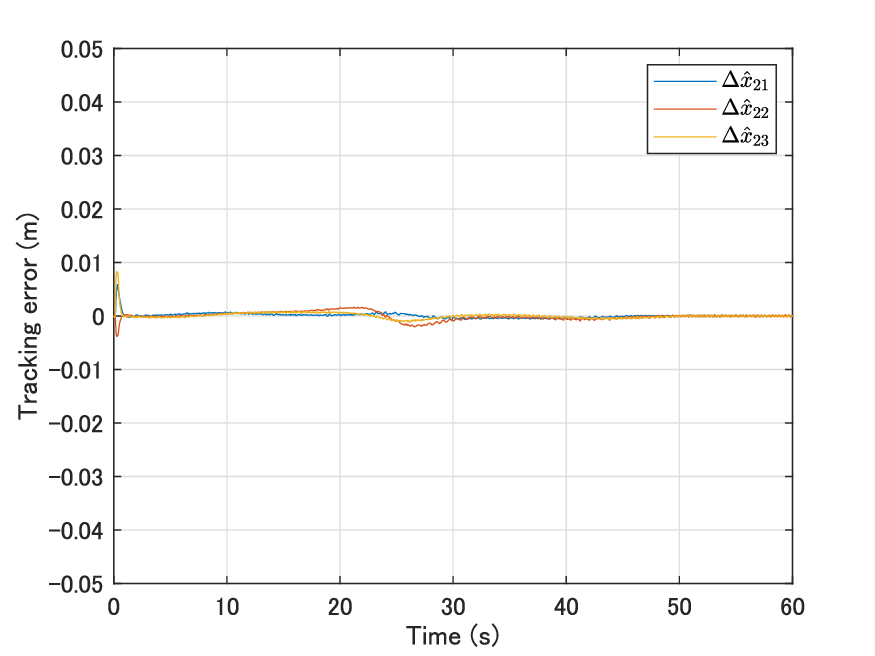}
        \caption{Estimated error for modular learning system 2}
        \label{fig:sub21}
    \end{subfigure}
    \vspace{1em}
    
    \begin{subfigure}[t]{0.47\textwidth}
        \centering
        \includegraphics[width=\textwidth]{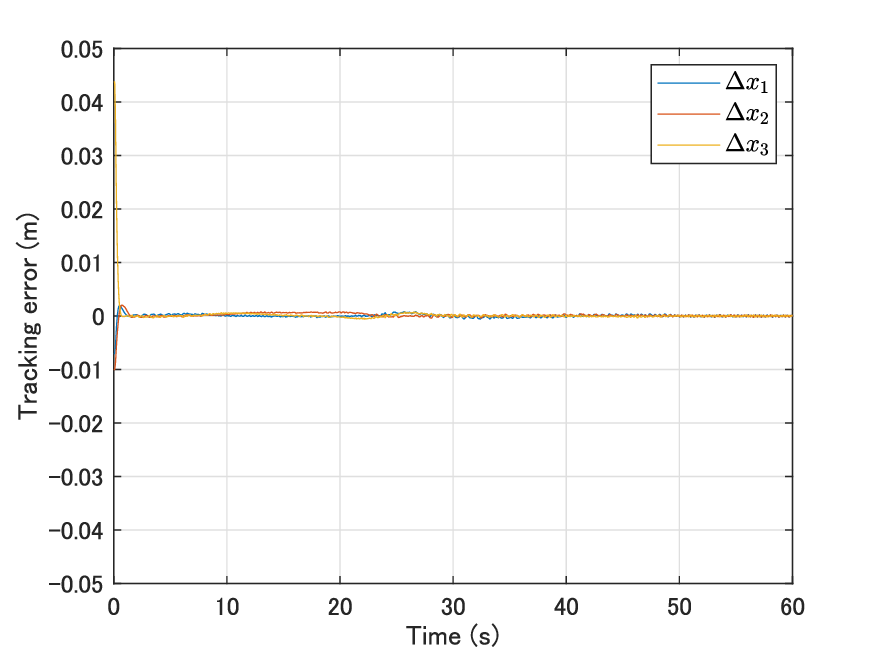}
        \caption{Tracking error}
        \label{fig:sub31}
    \end{subfigure}
    \hfill
    \begin{subfigure}[t]{0.47\textwidth}
        \centering
        \includegraphics[width=\textwidth]{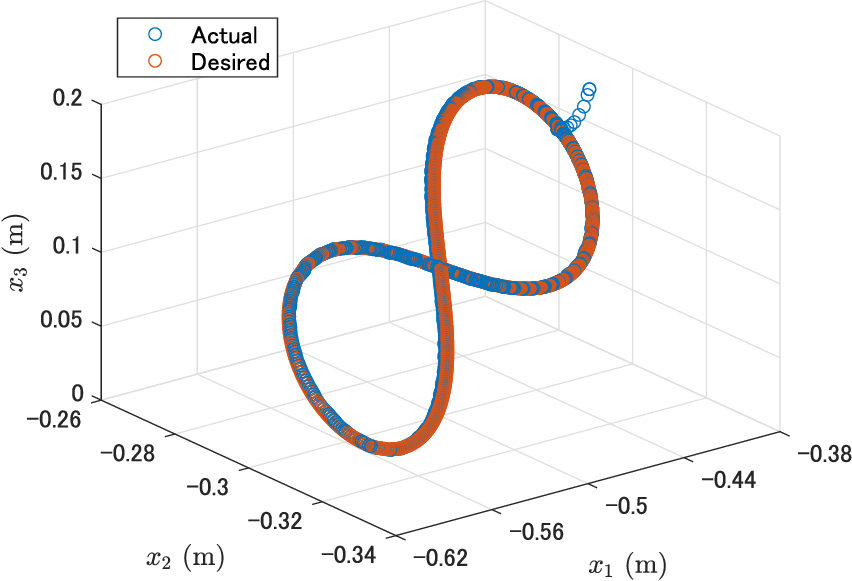}
        \caption{Trajectory tracking in 3D space}
        \label{fig:sub41}
    \end{subfigure}
    
    \caption{Results of tracking control}
    \label{fig:main3}
\end{figure*}

\hspace*{1em}To illustrate the performance of the proposed deep learning-based kinematic control with unknown Jacobian matrix, the experiments were conducted using an industrial robot, the UR5e, manufactured by Universal Robots. The proposed approach employs deep neural networks with three hidden layers, structured as 3-24-24-24-3. The three inputs are the three joint angles which form the major axes of the robot. Each hidden layer contains 24 neurons with sigmoid activation functions in each layer. The output layer comprises three neurons that generate the task-space velocities. \\
\hspace*{1em}In real-time task-space robot control, the Jacobian matrix plays an important role of providing directional information in transforming the task-space errors to joint-space commands. To enhance performance, the weights of the neural networks were pre-trained offline using a dataset collected around a home position of the robot. This home position $(\bm x = [0.2672, -0.5157, 0.5309]^\textrm T \ \textrm m)$ was set as the reference point for the entire experiment. The training dataset includes joint angles and joint velocities along with the corresponding end-effector velocities, which were collected from a region around the home position. This pre-training allows the network to capture the kinematic characteristics of the robot, contributing to improved stability and performance during online control. \\
\hspace*{1em}To perform the real-time control experiments, the target trajectory is defined as the lemniscate of Bernoulli in three-dimensional space, located far from the home region which is not covered by the dataset. The robot is programmed to follow this path, described by
\begin{align}
\bm{x}_d\left(t\right) = \bm{c} + \frac{\cos\left(\theta\left(t\right)\right)}{1 + \sin^2\left(\theta\left(t\right)\right)}\left(\bm r_1 + \sin\left(\theta\left(t\right)\right)\bm r_2\right) \ \mathrm{m}, \notag 
\end{align}
\begin{align}
\theta\left(t\right) = 2\pi\left(-2\left(\frac{t}{T}\right)^3 + 3\left(\frac{t}{T}\right)^2\right) \ \mathrm{rad} \notag
\end{align}
where $\bm c = [-0.5, -0.3, -0.1]^\textrm T \ \textrm m$, $\bm{r}_1 = [0.1, 0.01, 0.05]^\textrm T \ \textrm m$, and $\bm{r}_2 = [0.01, 0.05, 0.1]^\textrm T \ \textrm m$, $t$ and $T$ denote the current time and end time respectively. The robot was first moved from the home region to the initial position $(\bm x = [-0.4070, -0.3000, 0.1940]^\textrm T \ \textrm m)$ closer to the desired trajectory. During this process, online weight updates are employed to ensure consistent control performance, even in regions outside the dataset. \\
\hspace*{1em}For the region control task, the weights are updated in real time according to equations $(\ref{eq9})$ through $(\ref{eq261})$ with the learning rate $\eta = 0.01$ and the parameters set to $\alpha_l = \alpha^o_l = \alpha_{n-1} = \alpha_n = 1$, $\beta_l = \beta^o_l = \beta_{n-1} = \beta_n = 1$ for $l = 1, 2$ and $ n = 4$. The other parameters are configured as $k_{pl} = k_p = \left(k_T - k_0\right)\left(-2\left(\frac{t}{T}\right)^3 + 3\left(\frac{t}{T}\right)^2\right) + k_0$ for $t < T$ and $k_{pl} = k_p = k_T$ for $t \geq T$ with the initial gain $k_0 = 0.01$ and the final gain $k_T = 10$, $b_{li} = b_{i} = 0.001$, $k_{sl} = k_s = 0.001$, $a_i = 0.001$ for $i = 1, 2, 3$, corresponding to equations $(\ref{eq16})$ and $(\ref{eq1})$ for each respective $l$. The experimental results are presented in Fig. $\ref{fig:main2}$, illustrating the convergence of tracking error towards the desired bounds. As seen, the tracking error steadily decreases over time and approaches the bounds. \\
\hspace*{1em}In the tracking control task, the parameters are set as $\eta = 0.01$, $\alpha_l = \alpha^o_l = \alpha_{n-1} = \alpha_n = 1$, $\beta_l = \beta^o_l = \beta_{n-1} = \beta_n = 1$ in equations $(\ref{eq9a})$ to $(\ref{eq261a})$. Additionally, $k_{pl} = k_p = 10$, $k_{sl} = k_s = 0.001$ are specified by equations $(\ref{eq16a})$ and $(\ref{eq1a})$ with the difference from region control being that $k_{pl}$ and $k_p$ are fixed at 10 in tracking control. The experimental results in Fig. $\ref{fig:main3}$ shows a significant reduction in tracking error, attributed to the removal of the bounds present in region control. 
\section{Conclusion}
\hspace*{1em}This paper presented a theoretical framework of end-to-end deep learning control that seamlessly integrates with existing robot control theories for unknown kinematics. By employing the Lyapunov-like method, our approach ensures theoretical stability in real-time robot control while concurrently enabling online kinematic estimation. The experimental results indicate that the proposed method provides reliable and stable tracking performance, demonstrating its potential effectiveness. Furthermore, this work addresses the black-box problem of deep learning by introducing a structured framework that enhances both the stability and convergence of robot systems. These findings confirm the practical applicability of the proposed method in real-world robotic systems, supported by both theoretical analysis and experimental validation. 



\bibliographystyle{plain}        
\bibliography{autosam}           



\end{document}